\documentclass[twoside,leqno,twocolumn]{article}

\usepackage[letterpaper]{geometry}

\usepackage{ltexpprt}
\usepackage[colorlinks]{hyperref}
\usepackage{epsfig}
\usepackage{graphicx}
\usepackage{amsmath}
\usepackage{amssymb}
\usepackage{multirow}
\usepackage{colortbl}

\graphicspath{{figs/}}

\newcommand{\eg}{\textit{e.g.}}
\newcommand{\ie}{\textit{i.e.}}
\newcommand{\etal}{\textit{et al.}}

\DeclareMathOperator*{\softmax}{softmax}
\DeclareMathOperator*{\expect}{\mathbb{E}}
\newcommand{\collection}[1]{\texttt{#1}}
\newcommand{\starcenter}[1]{$\star$ #1 \hphantom{1}}

\newcommand{\todoref}[1]{\textcolor{red}{TODO}}

\begin{document}
% ========================================================================= %
%                           TITLE AND STUFF
% ========================================================================= %
\title{\Large Distillation from heterogeneous unlabeled collections}
\author{Jean-Michel Begon\thanks{\href{mailto:jm.begon@uliege.be}{jm.begon@uliege.be}; University of Liege.}
\and Pierre Geurts\thanks{\href{mailto:p.geurts@uliege.be}{p.geurts@uliege.be}; University of Liege.}}
\date{}

\maketitle

% Default Copyright Statement
%\fancyfoot[R]{\scriptsize{Copyright \textcopyright\ 2022 by SIAM\\
%Unauthorized reproduction of this article is prohibited}}

\begin{abstract} \small\baselineskip=9pt
Compressing deep networks is essential to expand their range of applications to constrained settings. The need for compression however often arises long after the model was trained, when the original data might no longer be available. On the other hand, unlabeled data, not necessarily related to the target task, is usually plentiful, especially in image classification tasks. In this work, we propose a scheme to leverage such samples to distill the knowledge learned by a large teacher network to a smaller student. The proposed technique relies on (i) preferentially sampling datapoints that appear related, and (ii) taking better advantage of the learning signal. We show that the former speeds up the student's convergence, while the latter boosts its performance, achieving performances closed to what can be expected with the original data.

% Max 6 KW
\end{abstract}

\section{Deep learning compression}
Compressing neural network offers the hope of leveraging the astounding performances of deep learning in constrained settings, thus offering a world of endless possibilities on  end-devices, far from huge computing servers and free of communication costs (and the associated privacy issues). Additional advantages include faster inference time, better energy efficiency, and possibly reduced overfitting.

Many approaches have been proposed to achieve this dream over the (last thirty) years, forming a few families, such as pruning (\eg{} \cite{lottery-ticket,brain-surgeon}), quantization (\eg{} \cite{zeroQ,quantized1}) or more simply designing (and searching) for small-yet-adequate architectures (\eg{} \cite{mobilenetv2,shufflenet,nasnet}). All these have in common the need for data, either to learn a model from scratch or process/fine-tune it. Those data, however, might no longer be available when the need for a compressed model arises. Maybe the data was lost, too large to store,  privacy prevent it or the network was manufactured and shipped ``as is''.

Orthogonal to those is distillation \cite{caruana-model-comp,hinto-ndistill},  where an already-learned model (the teacher), supposed to have captured the statistical information buried in the data, is available to ``teach'' another network (the student). Arguably, the gain from this teaching (a.k.a. distillation) compared to simply using the original dataset is small past some regularization \cite{label-smoothing}, and some speed up. Distillation shines, however, when there is no/few/more data.

\paragraph{Goal and scope.}
In this paper, we tackle the task of compressing a large neural network into a smaller one without the original dataset. We suppose the availability of (i) a teacher network which has been learned and performs well on a target task, and (ii) an unlabeled collection containing ``relevant'' data. 

Solutions which fit our setting have been proposed (see Section \ref{sec:related-works}) but come with a large additional computation cost compared to what would be required to learn the small network directly, were the data available. We would like to leverage the unlabeled collection of data to propose a (much) faster alternative.

We restrain our work to image classification where the availability of unlabeled samples might not potentially consist a hindrance. We focus here on students having potentially a totally different architecture as the teacher (\ie{} not necessarily a quantized or pruned version) since it is the more general setting.

\paragraph{Contribution.}
In light of these, our contributions can be summarized as follows:

\begin{enumerate}
    \setlength\itemsep{-.1em}
    \item we propose a method to focus on relevant samples from a collection of unlabeled data;
    \item we propose a fast solution to tackle distillation when original data is missing but such a collection is available;
    \item we conduct an extensive empirical study of the proposed solution and show how and when it is called for.
\end{enumerate}

\paragraph{Outline.}
We first formalize the problem we are tackling in Section \ref{sec:formalizaiton}, and discuss related works in Section \ref{sec:related-works}. Section \ref{sec:sampling-collection} exposes our two-part solution: preferential sampling (Section \ref{sec:biased-sampling}) and fully taking advantage of the learning signal (Section \ref{sec:fix-latent-space}). 
Our empirical study is presented in Section \ref{sec:empirical}. After a description of the protocol, we analyze the relevance of the collection (Section \ref{sec:main-results}), the impact of the biasing mechanism (Section \ref{sec:result-sampling}), and discuss the appropriateness of our distillation scheme (Section \ref{sec:ablation}), as well as broach a few other topics (Section \ref{sec:misc}). Section \ref{sec:conclusion} concludes.

\subsection{Formalization}\label{sec:formalizaiton}
Let $\mathcal{I}$ be the distribution of interest (also called the target or original distribution) on which a neural network $\hat{y}_t (\cdot, \Psi): \mathcal{X} \rightarrow \mathbb{R}^K$ was learned such that
\begin{align}
    \Psi \approx \min_{\Psi'} \expect_{x,y \sim \mathcal{I}} \Big\{ \ell \big(\hat{y}_t(x;\Psi'),y \big) + \mu R(\Psi')\Big\}
\end{align}
where $\mu$ weighs the two components of the loss, the regularization function $R$ is typically the weight decay and the loss function $\ell$ is usually the cross-entropy on the softmax logits (\ie{} outputs) in the case of classification (with $K$ classes):
\begin{align}
    \ell \big(\hat{y}_t(x;\Psi),y \big) &= -\sum_{j=1}^K y^{(j)} \log \hat{p}_{j}(x; t, \Psi) \\
    \hat{p}_j (x; t, \Psi) &= \frac{e^{\hat{y}_t^{(j)}(x;\Psi)}}{\sum_{k=1}^K e^{\hat{y}_t^{(k)}(x;\Psi)}} = \softmax_j \left( \hat{y}_t (x ; \Psi) \right)
\end{align}

Our goal is to learn a student network parametrized by $\Theta$ (rather than $\Psi$), denoted $\hat{y}_s (\cdot, \Theta): \mathcal{X} \rightarrow \mathbb{R^K}$, so that
\begin{align}
    \Theta \approx \min_{\Theta'}  \expect_{x\sim \mathcal{I}} \Big\{ \ell \big(\hat{y}_s(x; \Theta'), \hat{y}_t (x;\Psi) \big) \Big\} \label{eq:learning-problem}
\end{align}
To do so, we have at our disposal an unlabeled collection of samples $\mathcal{C} = \{x_i \in \mathcal{X}\}_{i=1}^n$, where $x_i \sim \mathcal{O}$ ($\neq \mathcal{I}$).

\subsection{Related works}\label{sec:related-works}
In semi-supervised distillation (\eg{} \cite{semi-sup-distill2,semi-sup-distill}), data scarcity is tackled by relying, in addition to some small learning sample $LS \sim \mathcal{I}^{m}$, on some unlabeled collection $\mathcal{C}'$. These methods leverage the capacity of the teacher to produce ``ground truth'' for the unlabeled samples, which are assumed to be drawn from the original distribution $\mathcal{I}$.

As early as in 2006, Buciluǎ \etal{} addressed data scarcity by using some form of data augmentation when training the student network \cite{caruana-model-comp}. Since then, more modern data-augmentation techniques have been proposed (\eg{} \cite{few-shot-gp,kdgan}). Another way to compensate for the lack of data is to extract more information from the teacher than only the output activations. For instance, one can exploit gradient information related to the decision boundary \cite{fleuret-jacobian} or use attention mechanism to teach more efficiently \cite{attention-few-shot2,attention-few-shot}. All those methods no longer need  $\mathcal{C}'$ but still assume the availability of some (potentially small) learning set $LS$ drawn from the original distribution. 

Recently, there has been a body of work trying to tackle the task of zero-short distillation: transferring the teacher knowledge without any data (from the original distribution). These methods share a common principle: derive from the teacher some out-of-distribution (OOD) \cite{ood} loss useful to craft informative samples. They differ on the actual loss which is used, as well as on how they craft these samples. For the latter, the two main investigated  techniques are (i) adversarially perturbing pure noise samples to minimize the OOD loss \cite{zeroQ,zs-adv2,zs-adv}, and (ii) including a generative adversarial network \cite{gan} in the loop \cite{zs-gan,zs-gan3,zs-gan2}.
Although offering good performance and needing neither $LS$ nor $\mathcal{C}'$, these methods are very heavy computation-wise, requiring thousands of adversarial perturbations or learning a whole generator network. Both approaches also introduce new hyper-parameters, including the whole generator architecture in the case of GAN, which are very difficult to tune in a setting where no data is available.% (arguably using GAN architectures that are known to work well for the distribution being transferred slightly bias the results).
% TODO be more quantitative on the cost of crafting examples?

Rather than manufacturing the inputs at great cost, we would like in this paper to take advantage of the availability of a collection of unlabeled images. The closest to this setting is the work in \cite{positive-unlabeled}, where the authors similarly train the student network on unlabeled images drawn from a distribution $\mathcal{O}\neq \mathcal{I}$. They however assume the availability of a small set of samples from $\mathcal{I}$, from which the most relevant samples from the collection are determined through a lengthy and sophisticated procedure.
%, which brings us close to the work of Xu \etal{}, where such a collection drawn from $\mathcal{O} \neq \mathcal{I}$ is available (from the cloud) but a small $LS$ set is still needed to select, through a lengthy and sophisticated technique, relevant images \cite{positive-unlabeled}.

\section{Distilling from an unlabeled collection}\label{sec:sampling-collection}
In this section, we introduce the two key components of our solution to tackle fast distillation from an unlabeled collection. Section \ref{sec:biased-sampling} presents how we propose to bias the sampling to rely mostly on informative samples, while Section \ref{sec:fix-latent-space} describes how to take full advantage of the learning signal. Note that we follow the notations of section \ref{sec:formalizaiton}.

\subsection{Biased sampling}\label{sec:biased-sampling}
In order to learn the student network, the loss in Eq. \ref{eq:learning-problem} must be estimated. It can be rewritten as
\begin{align}
  \expect_{x \sim \mathcal{I}} \Big\{  \ell \big( \hat{y}_s(x), \hat{y}_t (x) \big) \Big\} &= \expect_{x \sim \mathcal{O}} \Big\{ \beta(x) \ell \big(\hat{y}_s(x), \hat{y}_t (x) \big) \Big \} \label{eq:reweighing} \\
  \beta(x) &\triangleq \frac{\rho_{\mathcal{I}(x)}}{\rho_{\mathcal{O}(x)}}
\end{align}
where $\beta(x)$ is the density ratio with $\rho_{\mathcal{I}}(x)$ (resp. $\rho_{\mathcal{O}}(x)$) the density of $x$ for distribution $\mathcal{I}$ (resp. $\mathcal{O}$). Reweighing the samples in the estimate is legitimate since they are supposed to be in the support of $\mathcal{O}$. If $\beta$ were known, one could train the network by uniformly sampling examples from the collection $\mathcal{C}$ and using Eq. \ref{eq:reweighing} as the training loss. We propose instead to bias the sampling of the datapoints from $\mathcal{C}$ proportionally to the $\beta(x)$ weights for efficiency reasons. This nevertheless requires to estimate the density ratio.

%Since we have the collection $\mathcal{C}$, we propose to use its elements for the transfer from teacher to student. Rather than sampling uniformly or going over all the samples at each epoch, we will bias the the sampling of the datapoints to reflect the density ratio $\beta$. This, however, is only feasible if we know this ratio.

\paragraph{Estimating the density ratio.}
Let us introduce a random binary variable $s$, with $s = i$ if $x$ is
drawn from $\mathcal{I}$ and $s=o$ if $x$ is drawn from
$\mathcal{O}$. We can rewrite the density ratio as (using Bayes' rule):
\begin{align}
     \beta(x) &= \frac{\rho_{\mathcal{I}}(x)}{\rho_{\mathcal{O}}(x)} = \frac{\rho (x | s = i)}{\rho (x | s = o)} = \frac{p(s = o)}{p(s=i)} \, \frac{p(s=i | x)}{p(s=o|x)} \\
     &= \frac{1-\pi}{\pi} \, \frac{p(s=i | x)}{1 - p(s=i|x)} \\
     &= \frac{1-\pi}{\pi} e^{\lambda u(x)}
     \label{eq:s-density-ratio}
\end{align}
with
\begin{align}
    u(x) &\triangleq \frac{1}{\lambda}  \log    \frac{p (s=i|x)}{1 -  {p (s=i|x)}}
\end{align}
being proportional to the log odds ratio of being from $\mathcal{I}$ given $x$.

Assuming $u(x)$ is known (see below), let us discuss how this can be turned into sampling probabilities.

\paragraph{From odd ratio to sampling.}
Since our goal is to sample from $\mathcal{C}$ with probabilities proportional to $\beta$, we can simply normalize them:
\begin{align}
    q_i &= \frac{\beta(x_i)}{\sum_{x \in \mathcal{C}} \beta(x)} = \frac{\frac{1-\pi}{\pi} e^{\lambda u(x_i)}}{\frac{1-\pi}{\pi} \sum_{x \in \mathcal{C}} e^{\lambda u(x)}} \\
    &= \softmax(\lambda u(x_i)) \label{eq:sampling-prob}
\end{align}
We will denote by $\mathcal{Q}(\mathcal{C}, u, \lambda)$ the distribution over $\mathcal{C}$ reflecting those probabilities (dropping the arguments when the context is clear). Note that $\pi$ needs not be estimated as far as the sampling probabilities are concerned.

\paragraph{Characterizing score.}
In practice, we do not have direct access to the log odd ratio $u(x)$. We, therefore, propose to turn to some data-free proxy $\hat{u}(x)$, henceforth denoted as characterizing score. 

Note that our design allows for some robustness when it comes to a bad choice of characterizing score. On the one hand, $\hat{u}$ needs only be a linear transformation of a good approximation of $u$ since the softmax is translation invariant and $\lambda$ can accommodate a scaling factor. On the other hand, even an imperfect choice of score might prove to be useful since we are not selecting datapoints but \emph{sampling} them.

In this paper, we will focus on two out-of-distribution (OOD) characterizing score. The first one, T1000, is based on the so-called ODIN method \cite{odin} and consists in running the logits through a softmax with a temperature $\tau=1000$, and taking the highest value. This is equivalent to a linear transformation of the highest logit. The second score, 1C-Sum \cite{1C-Sum}, is a combination of several other OOD indicators, among which the entropy, T1000, the cosine distance between the latent vector and the direction of the hyperplane with highest logit and batch-normalization-based features.

\paragraph{Controlling the diversity.} To train our student, we will sample examples in each training batch from $\mathcal{C}$ using distribution $\mathcal{Q}$. To ensure some diversity in the selected samples and provide some robustness against an inappropriate choice of characterizing score (as mentioned above), we can adjust the $\lambda$ parameter, which controls an exploitation-exploration tradeoff. 
If $\lambda=0$, the distribution is uniform, while as $\lambda \to \infty$ only the sample(s) appearing the most to come from $\mathcal{I}$ will be selected.

Optimizing the hyper-parameter $\lambda$ cannot be done in the usual fashion in our setting but choosing a value can be done in many ways. Here we propose an intuitive, non-parametric solution. Let $l$ (resp. $h$) be the index of the $l$th (resp. $h$th) sample of $\mathcal{C}$ in ascending order of value of $\hat{u}$, by choosing a value for $q_h / q_l$ we can isolate $\lambda$ from
\begin{align}
    \log \frac{q_h}{q_l} = \lambda (\hat{u}(x_h) - \hat{u}(x_l)). %\iff \lambda = \frac{1}{\hat{u}(x_h) - \hat{u}(x_l)} \log \frac{q_h}{q_l}
\end{align}
We propose to select $l$ (resp. $h$) to correspond to the first (resp. third) quartile and dub $q_h / q_l$ the inter-quartile sampling probability ratio (IQPR). If $\text{IQPR}=5$, for instance, the sample corresponding to the third quartile has five time more chance of being selected than the sample at the first one. Note that when $\text{IQPR}=1$, $\mathcal{Q}$ is uniform.

\paragraph{On computational cost.}
Biasing the sampling incurs an additional cost of one forward pass of the teacher per sample in the collection in order to compute $\hat{u}(x)$. This is admittedly much less than what is required to learn a good generator or adversarially transform pure noise into useful samples, as proposed in the context of zero-shot distillation (see Section \ref{sec:related-works}).
Note that we could have sampled mini-batches uniformly from $\mathcal{C}$ and then reweighed them in the loss as in Eq. \ref{eq:objective} (like in \cite{few-shot-gp,semi-sup-distill}). We believe, however, that there is no need to spend too much time on uninformative samples.

\subsection{Capturing the learning signal: fixed-linear distillation} \label{sec:fix-latent-space}

Traditional teacher-student transfer encourages the student to replicate the teacher outputs, either by imposing a $L_2$ norm on the logit \cite{caruana-model-comp} or, more frequently, a (softened) cross entropy loss on the output probabilities \cite{hinto-ndistill}. The latter will be referred to as classical (or vanilla) distillation, and the corresponding loss takes the form
\begin{align}
  \ell_{t, s, \tau}(x) &= - \sum_j^K \tilde{p}_{j, \tau} (x, t) \log \tilde{p}_{j, \tau} (x, s),\\
  \mbox{with }  \tilde{p}_{j, \tau} (x, \centerdot) &= \softmax_j \left( \frac{1}{\tau} \hat{y}_{\centerdot} (x) \right),
\end{align}
where the temperature $\tau$ is the hyper-parameter responsible for softening the probability distribution (the higher $\tau$, the more uniform the distribution).

However, many problems (including most practical ones) have a small number of outputs. As a consequence, trying to replicate only those is wasteful sample-wise; a better approach would be to extract more constraints per input from the teacher's inner functioning.
Attention mechanisms, whose goal is to force other parts of the student to behave as the teacher, have been proposed to guide and accelerate student training (\eg{} \cite{attention-gan, attention-few-shot, bengio-fitnet, attention2,attention1}). Although well motivated when inner parts of the student and teacher networks can be matched, they can not be applied in our more general setting where the two architectures are not (necessarily) related.

We propose to take advantage of the only part where a mapping can be expected even for unrelated architectures, \ie{}, at the end of the feature extraction phase. Indeed, most architectures can be decomposed as:
\begin{align}
    \hat{y}(x; \Theta) = W z(x; \theta)  + b
\end{align}
where $z(\cdot; \theta) : \mathcal{X} \rightarrow \mathbb{R}^p$ is a feature extractor and  $\Theta = [\theta, W, b]$. Our idea is to enforce the student to match the teacher's feature representation. Since $p\gg K$ (remember that $K$ is the number of classes and $p$ is the dimensionality of the latent space), this will put more constraints during training. By doing that, we hope to be  more efficient when learning from samples which are not from the target distribution.

To formalize this idea, let us denote by $z_s \in \mathbb{R}^{p_s}$ (resp. $z_t \in \mathbb{R}^{p_t}$) the student (resp. teacher) latent vector corresponding to some $x$, and use the corresponding subscript for $W, b$. Given that $p_s$ might be different from $p_t$, the learning problem is then defined as follows:
\begin{align}
\begin{cases}
    &W_s = P W_t \\
    &b_s = b_t \\
    &\min_{\theta, P} \expect_{x \sim \mathcal{Q}} || P z_s(x; \theta) - z_t(x; \psi) ||_2^2
\end{cases}\label{eq:objective}
\end{align}
that is, we fix the linear part of the student and learn the same feature extraction as the teacher's. $P$ serves to project the student latent vector onto the teacher space when $p_S \neq p_t$ so that $W_t z_t = W_s z_s = W_t P z_s$. When $p_s = p_t$, one should take $P$ as the identity matrix and drop it from the objective. In contrast to how traditional attention mechanisms are incorporated in the learning, the loss has only one component and therefore no weighing hyper-parameter needs to be tuned (which would be hard to do in our setting).

Although it is possible to fine-tune the linear part of the student by classical distillation, we expect the gain to be small (that part has already been optimized on the teacher) compared to the risk of disrupting the model (using a poorly chosen fine-tuning learning rate, for instance). 

We will refer to Eq. \ref{eq:objective} as fixed-linear distillation.

\section{Empirical analysis}\label{sec:empirical}

% TODO architecture, protocol, dataset, etc.
In this section, we analyze the role played by the collection (Section \ref{sec:main-results}), followed by a discussion of the influence of biasing the sampling (Section \ref{sec:result-sampling}) and a study of fixed-linear distillation (Section \ref{sec:ablation}). We then cover a few additional questions naturally raised by our methods (Section \ref{sec:misc}). We start by describing our protocol.

\subsection{Protocol}

\paragraph{Tasks.}
We evaluate our methodology extensively on CIFAR 10 \cite{cifar} %, ImageNet \cite{imagenet} 
and more briefly on KMNIST \cite{kmnist}, using their standard test sets to assess model performance.

To constitute the collection of unlabeled images, we used MNIST \cite{mnist}, Fashion MNIST \cite{fashionmnist}, SVHN \cite{svhn}, STL 10 \cite{stl10} (with all the unlabeled images) and tiny ImageNet \cite{tinyimagenet}. We used both train and test sets for the collection, only keeping $10\%$ of the train set as validation set to monitor the loss. When the collection is made up of several datasets, we concatenated those validation sets. When images from the original task appear in the collection, images from the test set are not included obviously. All images were resized and cast to RGB to fit the network expectations.

We grouped some datasets to form meaningful collections as shown in Table \ref{tab:collections}. \collection{Ori} is simply the original task and is included for the sake of the discussion. \collection{Rel} stands for relevant, \ie{} datasets whose label-space intersects with the original task's. \collection{Irrel} contains only irrelevant datasets. We will also consider three combinations of those collections. The second part of Table \ref{tab:collections} holds the relative size of each sub-collection. Note that the most realistic collection is \collection{rel + irrel}.

\paragraph{Teachers.}
ResNet 50 \cite{resnet} and DenseNet 121 \cite{densenet} were used as teachers. All networks expect RGB images of size $224 \times 224$. %For ImageNet, we re-used the available weights offered by PyTorch \cite{pytorch}, thus only one run is reported in the experiments. On CIFAR 10, 
The networks were optimized during $450$ epochs by stochastic gradient descent (batches of size $64$, weight decay of $5\times 10^{-4}$ , and momentum of $0.9$). The learning rate was initialized at $0.1$. It was decreased by a factor $10$ after $150$ epochs and again at epoch $300$. Each decrease was accompanied by a restart from the best model according to the validation accuracy. Horizontal flip and random cropping (with padding of $4$) were used as data augmentation. 
Teacher accuracies are shown in Table \ref{tab:teacher-acc}.

%\begin{table}[t]
%	%\vspace{-1em}
%	\caption{Test set accuracy (in $\%$) of the teacher networks. R50 and D121 stands for ResNet 50 and DenseNet 121 respectively. \label{tab:teacher-acc}}
%	%\vspace{.5em}
%	\centering
%	\small
%	\begin{tabular}{c|c|c|c}
%		\hline
%		& CIFAR 10 & CIFAR 100 & KMNIST \\
%		\hline
%		R50 & 94.11 $\pm$ 0.25 & 77.48 $\pm$ 0.23  & 98.85 $\pm$ 0.01 \\
%		D121 & 94.30 $\pm$ 0.31 & 77.89 $\pm$ 0.04 & -  \\
%		\hline
%	\end{tabular}
%\end{table}

\paragraph{Students and distillation.}
We used MobileNet v2 \cite{mobilenetv2} and ShuffleNet v2 \cite{shufflenetv2} as students, which expect RGB images of size $224 \times 224$ as well. The students were optimized with vanilla distillation or according to Eq. \ref{eq:objective} with different values of IQPR. They were trained for the equivalent of $150$ epochs of the target task's training set so as to reach convergence. To reach fast convergence, we considered pseudo-epochs of $5000$ samples (accounting for 1350 of such pseudo-epochs) drawn with replacement from the collection, and  divided the learning rate (initialized at $0.01$) by approximately $0.4$ after no improvement was seen on the held-out validation set from the collection for $20$ pseudo-epochs. We used the same data augmentation scheme as for the teachers (data augmentation is performed after bias sampling). The results are averaged over three random initializations of the teacher and student networks. Unless mentioned otherwise, experiment results are based on 1C-Sum with CIFAR 10 as target task. 
All experiments were carried out with PyTorch \cite{pytorch}.

\begin{table}[t]
    %\vspace{-1em}
  \caption{Details of the collections used as proxy. \label{tab:collections}}
  %\vspace{.5em}
    \centering
    \small
    \begin{tabular}{l|l|r|c}
    \hline
    Collection & Included & Transfer Size & Reference \\
    \hline
    \collection{Ori} & CIFAR 10 & 55000 & \cite{cifar} \\
    \hline
    \multirow{2}{*}{\collection{Rel}} & Tiny ImageNet & 100000 & \cite{tinyimagenet}\\
    & STL 10 & 102500 & \cite{stl10} \\
    \hline
    \multirow{3}{*}{\collection{Irrel}} & MNIST (x2) & 128000 & \cite{mnist}\\
    & Fashion MNIST & 64000  & \cite{fashionmnist}\\
    & SVHN & 91963 & \cite{svhn}\\
    \hline \hline
    \end{tabular}
    
    %\begin{tabular*}{.45\textwidth}{l|c|c|c}
    \begin{tabular}{l|c|c|c}
    & \multicolumn{3}{c}{Relative size (in \%)} \\
    Collection & \collection{Ori} & \collection{Rel} & \collection{Irrel} \\
    \hline
    \collection{Ori + rel} & 21.36 & 78.64 & - \\
    \collection{Ori + irrel} & 16.23 & - & 83.77 \\
    \collection{Rel + irrel} & - & 41.63 & 58.37 \\
    %\hline
    %\end{tabular*}
    \end{tabular}
    \hrule
    \vspace{-1em}
\end{table}

\begin{table}[t]
    %\vspace{-1em}
  \caption{Test set accuracy (in $\%$) of the teacher networks.\label{tab:teacher-acc}}
  %\vspace{.5em}
    \centering
    \small
    \begin{tabular}{c|c|c}
    \hline
        & CIFAR 10  & KMNIST \\
        \hline
        ResNet 50 & 94.11 $\pm$ 0.25 &  98.85 $\pm$ 0.01 \\
        DenseNet 121 & 94.30 $\pm$ 0.31 & -  \\
        \hline
    \end{tabular}
\end{table}

\subsection{Collection analysis}\label{sec:main-results}
Table \ref{tab:main-result-cifar} shows test set accuracy on CIFAR 10 for several from/to architecture pairs trained with fixed-linear distillation under different collections as transfer set and three IQRP values. We defer the comparison with vanilla distillation to Section \ref{sec:ablation}.

As a sanity check, let us note that the best performance (up to $94 \%$ of accuracy) is obtained by using the original dataset without biasing the sampling (last column), a setting which is supposed to be impossible in our context. Two more prominent observations can be made: highly biasing the sampling can be harmful and only using irrelevant data results in very poor accuracies (although one might be surprised at how high an accuracy is achievable with such unrelated data).

Excellent performances can usually be obtained when the original data is part of the transfer collection. Interestingly, we see that even in the worst-case situation (\collection{ori + irrel}, where good data represents less than one fifth of the collection) very decent accuracy can be reached with uniform sampling, although biasing the sampling is most useful in this situation. This suggests that student training is robust to such irrelevant data, probably because the network is not saturated.

When only relevant samples are available (\collection{rel}), biasing makes little sense (IQPR=1 offers the best accuracy). However, in the most realistic setting, where relevant and irrelevant samples from the collection (\collection{rel + irrel}), a small bias (IQPR=5) usually offers a slight edge.

\begin{table*}[t]
    %\vspace{-1em}
  \caption{CIFAR 10 test set accuracy (in $\%$) after distillation with respect to the collection used as transfer set. Mob. and Shuff. stands for MobileNet and ShuffleNet respectively. \label{tab:main-result-cifar}}
  %\vspace{.5em}
    \centering
    \small
    \begin{tabular}{cc|c||c|c|c|c|c||c}
    \hline
    From & To & IQPR&\collection{Rel + irrel}& \collection{Rel}&\collection{Irrel}&\collection{Ori + rel}& \collection{Ori + irrel}& \collection{Ori}\\%
    \hline 
        \multirow{6}{*}{\rotatebox[origin=c]{90}{ResNet 50}} & \multirow{3}{*}{\rotatebox[origin=c]{90}{Mob.}} & %
        1 & \cellcolor[rgb]{0.29, 0.59, 0.79} 91.46 $\pm$ 0.27 & \cellcolor[rgb]{0.29, 0.59, 0.79} 92.17 $\pm$ 0.08 & \cellcolor[rgb]{0.29, 0.59, 0.79} 72.24 $\pm$ 4.69 & \cellcolor[rgb]{0.29, 0.59, 0.79} 93.90 $\pm$ 0.21 & \cellcolor[rgb]{0.97, 0.98, 1.00} 93.06 $\pm$ 0.29 & \cellcolor[rgb]{0.29, 0.59, 0.79} 94.15 $\pm$ 0.34\\ 
        & & 5 & \cellcolor[rgb]{0.44, 0.69, 0.84} 91.04 $\pm$ 0.15 & \cellcolor[rgb]{0.34, 0.63, 0.81} 90.99 $\pm$ 0.35 & \cellcolor[rgb]{0.50, 0.73, 0.86} 68.67 $\pm$ 2.22 & \cellcolor[rgb]{0.40, 0.67, 0.83} 93.83 $\pm$ 0.55 & \cellcolor[rgb]{0.29, 0.59, 0.79} 93.99 $\pm$ 0.46 & \cellcolor[rgb]{0.36, 0.64, 0.81} 93.73 $\pm$ 0.23\\ 
        & & 25 & \cellcolor[rgb]{0.97, 0.98, 1.00} 89.25 $\pm$ 0.46 & \cellcolor[rgb]{0.97, 0.98, 1.00} 74.10 $\pm$ 9.15 & \cellcolor[rgb]{0.97, 0.98, 1.00} 58.17 $\pm$ 2.12 & \cellcolor[rgb]{0.97, 0.98, 1.00} 93.38 $\pm$ 0.28 & \cellcolor[rgb]{0.65, 0.81, 0.89} 93.60 $\pm$ 0.26 & \cellcolor[rgb]{0.97, 0.98, 1.00} 89.44 $\pm$ 0.27\\ 
        \cline{2-9}
        & \multirow{3}{*}{\rotatebox[origin=c]{90}{Shuff.}} & %
       1 & \cellcolor[rgb]{0.40, 0.74, 0.43} 90.34 $\pm$ 0.17 & \cellcolor[rgb]{0.29, 0.69, 0.38} 91.47 $\pm$ 0.05 &  \cellcolor[rgb]{0.29, 0.69, 0.38} 66.14 $\pm$ 2.25 &  \cellcolor[rgb]{0.32, 0.70, 0.40} 93.36 $\pm$ 0.31 &  \cellcolor[rgb]{0.97, 0.99, 0.96} 91.93 $\pm$ 0.20 &  \cellcolor[rgb]{0.29, 0.69, 0.38} 93.69 $\pm$ 0.25\\ 
        & & 5 & \cellcolor[rgb]{0.29, 0.69, 0.38} 90.63 $\pm$ 0.01 & \cellcolor[rgb]{0.38, 0.73, 0.42} 90.59 $\pm$ 0.00 & \cellcolor[rgb]{0.41, 0.75, 0.44} 64.78 $\pm$ 1.85 & \cellcolor[rgb]{0.29, 0.69, 0.38} 93.47 $\pm$ 0.40 & \cellcolor[rgb]{0.67, 0.87, 0.64} 92.37 $\pm$ 0.65 & \cellcolor[rgb]{0.46, 0.77, 0.47} 92.77 $\pm$ 0.08\\ 
        & & 25 & \cellcolor[rgb]{0.97, 0.99, 0.96} 87.90 $\pm$ 0.36 & \cellcolor[rgb]{0.97, 0.99, 0.96} 80.93 $\pm$ 0.71 & \cellcolor[rgb]{0.97, 0.99, 0.96} 55.06 $\pm$ 1.54 & \cellcolor[rgb]{0.97, 0.99, 0.96} 89.48 $\pm$ 3.10 & \cellcolor[rgb]{0.29, 0.69, 0.38} 92.69 $\pm$ 0.20 & \cellcolor[rgb]{0.97, 0.99, 0.96} 88.34 $\pm$ 0.09\\ 
        \hline \hline
        \multirow{6}{*}{\rotatebox[origin=c]{90}{DenseNet 121}} & \multirow{3}{*}{\rotatebox[origin=c]{90}{Mob.}} & %
        1 & \cellcolor[rgb]{1.00, 0.96, 0.92} 91.27 $\pm$ 0.40 & \cellcolor[rgb]{0.95, 0.44, 0.11} 91.60 $\pm$ 0.35 & \cellcolor[rgb]{0.95, 0.44, 0.11} 76.43 $\pm$ 0.80 & \cellcolor[rgb]{0.96, 0.45, 0.12} 93.92 $\pm$ 0.26 & \cellcolor[rgb]{1.00, 0.96, 0.92} 93.06 $\pm$ 0.17 & \cellcolor[rgb]{0.95, 0.44, 0.11} 94.36 $\pm$ 0.25\\ 
        & & 5 & \cellcolor[rgb]{0.95, 0.44, 0.11} 91.89 $\pm$ 0.04 & \cellcolor[rgb]{0.97, 0.48, 0.16} 91.41 $\pm$ 0.61 & \cellcolor[rgb]{0.99, 0.55, 0.23} 75.89 $\pm$ 0.96 & \cellcolor[rgb]{0.95, 0.44, 0.11} 94.05 $\pm$ 0.13 & \cellcolor[rgb]{0.95, 0.44, 0.11} 93.93 $\pm$ 0.25 & \cellcolor[rgb]{0.99, 0.53, 0.21} 93.81 $\pm$ 0.14\\ 
        & & 25 & \cellcolor[rgb]{0.99, 0.72, 0.48} 91.62 $\pm$ 0.49 & \cellcolor[rgb]{1.00, 0.96, 0.92} 88.85 $\pm$ 0.33 & \cellcolor[rgb]{1.00, 0.96, 0.92} 73.01 $\pm$ 0.02 & \cellcolor[rgb]{1.00, 0.96, 0.92} 87.68 $\pm$ 6.78 & \cellcolor[rgb]{0.99, 0.63, 0.35} 93.68 $\pm$ 0.78 & \cellcolor[rgb]{1.00, 0.96, 0.92} 90.29 $\pm$ 0.12\\ 
        \cline{2-9}
        & \multirow{3}{*}{\rotatebox[origin=c]{90}{Shuff.}} & %
        1 & \cellcolor[rgb]{0.99, 0.98, 0.99} 90.37 $\pm$ 0.01 & \cellcolor[rgb]{0.53, 0.51, 0.74} 91.50 $\pm$ 0.53 & \cellcolor[rgb]{0.53, 0.51, 0.74} 70.35 $\pm$ 0.59 & \cellcolor[rgb]{0.99, 0.98, 0.99} 93.33 $\pm$ 0.17 & \cellcolor[rgb]{0.99, 0.98, 0.99} 91.79 $\pm$ 0.20 & \cellcolor[rgb]{0.53, 0.51, 0.74} 93.71 $\pm$ 0.01\\ 
       & & 5 & \cellcolor[rgb]{0.53, 0.51, 0.74} 91.14 $\pm$ 0.16 & \cellcolor[rgb]{0.54, 0.53, 0.75} 91.24 $\pm$ 0.35 & \cellcolor[rgb]{0.53, 0.52, 0.74} 70.31 $\pm$ 1.09 & \cellcolor[rgb]{0.53, 0.51, 0.74} 93.83 $\pm$ 0.28 & \cellcolor[rgb]{0.58, 0.57, 0.77} 93.24 $\pm$ 0.15 & \cellcolor[rgb]{0.64, 0.62, 0.80} 92.51 $\pm$ 0.08\\ 
        & & 25 & \cellcolor[rgb]{0.89, 0.88, 0.94} 90.63 $\pm$ 0.11 & \cellcolor[rgb]{0.99, 0.98, 0.99} 83.46 $\pm$ 2.59 & \cellcolor[rgb]{0.99, 0.98, 0.99} 67.90 $\pm$ 1.89 & \cellcolor[rgb]{0.68, 0.67, 0.82} 93.70 $\pm$ 0.11 & \cellcolor[rgb]{0.53, 0.51, 0.74} 93.41 $\pm$ 0.02 & \cellcolor[rgb]{0.99, 0.98, 0.99} 87.67 $\pm$ 0.78\\ 
        \hline
    \end{tabular}
\end{table*}

\subsection{Sampling analysis}\label{sec:result-sampling}
\begin{figure}[t]
\begin{center}
\centerline{\includegraphics[width=\columnwidth]{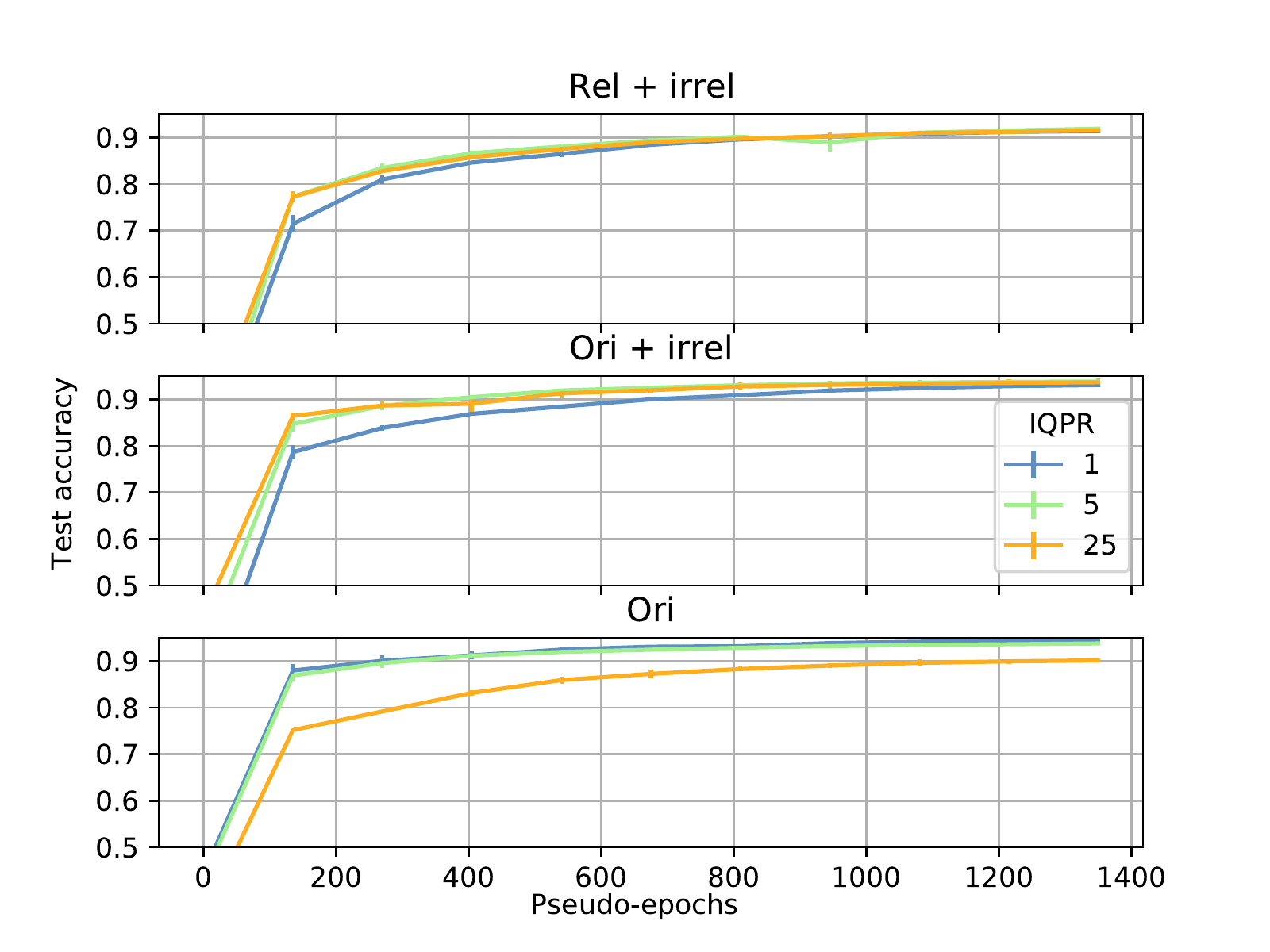}}
\caption{Convergence rate: CIFAR 10 test set accuracy with respect to the learning time and for several collections and IQPR values (based on DenseNet 121 to MobileNet).}
\label{fig:convergence}
\end{center}
\vspace{-2em}
\end{figure}

In this section, we investigate the effect of the biased sampling. Although the previous section suggested that biasing had little impact, we can see from Figure \ref{fig:convergence} that it tends to accelerate the convergence to the final accuracy. On \collection{rel + irrel}, using an IQPR of 5 instead of 1 results in an average accuracy of $77.5\%$ instead of $69.1\%$ at $10\%$ of the learning. The gap remains wide during the whole training on \collection{ori + irrel}. When using the original data (\collection{ori}), biasing downgrades the performance, by masking examples. We expect biasing to provide an advantage in this setting only if the dataset contains outliers.

%biasing would only provide an advantage if the dataset contained outliers. Otherwise, biasing acts like masking data and downgrades the performances. 
%Unfortunately, biasing also tends to somewhat increase instability during training.

Table \ref{tab:uniformity} offers more insight into the sampling mechanism. Skip ratio represents the percentage of samples from the collection which never get selected. Uniformity is the entropy of the empirical selection distribution. It is rescaled in the range $[0, 1]$ (close to $1$ means uniform, close to $0$ means highly biased) and ignores samples which are never selected. Finally, irrel. prop. is the percentage of the samples that are used at least once during the training that comes from \collection{irrel}.

When there is a slight bias (IQPR=5), only a small fraction of the data is ignored. This percentage is the smallest on \collection{ori} where the scores used to compute the actual sampling probabilities are supposed to be more uniform.
% TODO show this
The proportion of irrelevant samples is drastically reduced in the case of \collection{ori + irrel}. This suggests that we are able to select samples from the original data quite well and explains the good results in Table \ref{tab:main-result-cifar} and Figure \ref{fig:convergence}.

When the bias is more severe (IQPR=25), we see that a great proportion (20 to 40 \%) of the data are totally discarded. Even for the remaining samples, the selection departs largely from a uniform distribution. As a consequence, many datapoints are mostly ignored. It is clear that this strategy can only pay off when the collection is polluted with many easily-identified irrelevant samples. Since this is close to pre-selecting the samples, this scheme also rids us of the benefits of compensating a bad choice of characterizing score. In any case, this accounts for the bad performances on \collection{rel} and \collection{ori} (Table \ref{tab:main-result-cifar}).

\begin{table*}[t]
    %\vspace{-1em}
  \caption{Biased sampling related metrics: skip ratio is the percentage of samples which are never selected; uniformity is the empirical entropy of the selected sample distribution; irrelevant proportion is the percentage of samples from \collection{irrel} which are used at least once during training (based on transferring CIFAR 10 from DenseNet 121 to MobileNet). Cells marked with $\star$ have non-negligible standard deviation (see Supplementary A.1 for more details).  \label{tab:uniformity}}
  %\vspace{.5em}
    \centering
    \small
    \begin{tabular}{c||c|c|c|c|c|c|c|c}
    \hline
    & \multicolumn{3}{c|}{\collection{Rel + irrel}} & \multicolumn{3}{c|}{\collection{Ori + irrel}} & \multicolumn{2}{c}{\collection{Ori}} \\
   IQPR&Skip ratio&Uniformity&Irrel. prop.&Skip ratio&Uniformity&Irrel. prop.&Skip ratio&Uniformity\\%
    \hline
    1 & 0.00 & 1.00 & 58.36 & 0.00 & 1.00 & 86.32 & 0.00 & 1.00\\ 
    5 & 3.53 & 0.97 & 42.12 & 1.91 & 0.96 & 56.65 & 0.37 & 0.94\\ 
    25 & 24.99 & 0.95 & \starcenter{38.84} & \starcenter{40.09} & \starcenter{0.86} & \starcenter{17.32} & \starcenter{21.48} & \starcenter{0.60}\\ 
    \hline
    \end{tabular}
\end{table*}

\subsection{Fixed-linear distillation analysis}\label{sec:ablation}
To assess the impact of fixed-linear distillation, we compare it to classical distillation. Since a projection is involved for our variant, we carried out two tests: the first one with the default version of the student architecture and the second one with a modified student network where the
pre-linear latent space is updated to match the dimensionality of the teacher's (2048 for ResNet 50). To do so, we simply modify the number of feature maps produced by the last convolution layer of our MobileNet student. The last convolution is indeed followed by global average pooling, resulting in one latent feature per feature map. In this case, the size of the latent space is increased (from 1280 to 2048), resulting in more parameters for this variant of the student. The results are collected in Table \ref{tab:ablation}. Fixed-lin. + proj. corresponds to Eq. \ref{eq:learning-problem}. Fixed-lin. 2048 is our method on the modified student. Distill 2048 is vanilla distillation on the modified student, while Distill corresponds to vanilla distillation on the original student. We used a temperature of 2 for all vanilla distillations.

On the \collection{rel + irrel} collection, there is a clear incentive to use the fixed-linear distillation throughout the whole learning procedure and irrespective of the IQPR. The projection, on the other hand, plays an insignificant role.

On the original data (\collection{ori}), the usefulness of taking advantage of the latent space information clearly disappears, offering only a slight edge at the start. Interestingly, the projection is better able to keep up with the distillation than when the student's latent space is made to match the teacher's.

Overall, it appears more critical to exploit well the learning signal when using proxy data. The conclusion drawn regarding biased sampling seems to hold with classical distillation as well (\ie{} useful on proxy data containing relevant samples to speed up convergence). By the end of learning, the modified student never significantly outperforms the original student and even tends to underperform (or at least is less stable) on the original data. As a consequence, there is no incentive to add more parameters to match the teacher.

\begin{table*}[t]
    %\vspace{-1em}
  \caption{CIFAR 10 test set accuracy (in $\%$) at 10\% and 100\% of learning with different distillation methods and collections. Fixed-lin. + proj is our method; distill is the classical distillation; 2048 refers to modifying the student to match the latent space dimensionality of the teacher (based on ResNet 50 to MobileNet). \label{tab:ablation}}
  %\vspace{.5em}
    \centering
    \small
    \begin{tabular}{c|c||c|c|c|c}
    \hline
    & & \multicolumn{2}{c}{IQPR = 1} & \multicolumn{2}{|c}{IQPR = 5} \\
    Collection & Method & @ 10 \% &  @ 100 \% &  @ 10 \% &  @ 100 \% \\
    \hline
    \multirow{4}{*}{\collection{Rel + irrel}} & %
    Fixed-lin. + proj. & \cellcolor[rgb]{0.41, 0.68, 0.84} 70.33 $\pm$ 2.37 & \cellcolor[rgb]{0.29, 0.59, 0.79} 91.46 $\pm$ 0.27 & \cellcolor[rgb]{0.29, 0.59, 0.79} 75.77 $\pm$ 0.07 & \cellcolor[rgb]{0.34, 0.63, 0.81} 91.04 $\pm$ 0.15\\ 
    & Fixed-lin. 2048 & \cellcolor[rgb]{0.29, 0.59, 0.79} 71.59 $\pm$ 2.65 & \cellcolor[rgb]{0.33, 0.62, 0.80} 91.34 $\pm$ 0.07 & \cellcolor[rgb]{0.33, 0.62, 0.80} 75.42 $\pm$ 0.74 & \cellcolor[rgb]{0.29, 0.59, 0.79} 91.18 $\pm$ 0.00\\ 
    & Distill 2048 & \cellcolor[rgb]{0.97, 0.98, 1.00} 63.52 $\pm$ 0.38 & \cellcolor[rgb]{0.90, 0.94, 0.98} 89.59 $\pm$ 0.46 & \cellcolor[rgb]{0.97, 0.98, 1.00} 68.45 $\pm$ 0.60 & \cellcolor[rgb]{0.95, 0.97, 1.00} 89.11 $\pm$ 0.38\\ 
    & Distill & \cellcolor[rgb]{0.94, 0.97, 0.99} 64.02 $\pm$ 2.02 & \cellcolor[rgb]{0.97, 0.98, 1.00} 89.26 $\pm$ 0.97 & \cellcolor[rgb]{0.93, 0.96, 0.99} 69.03 $\pm$ 3.22 & \cellcolor[rgb]{0.97, 0.98, 1.00} 89.04 $\pm$ 0.75\\ 
    \hline \hline
    \multirow{4}{*}{\collection{Ori}} & %
    Fixed-lin. + proj. & \cellcolor[rgb]{0.99, 0.85, 0.71} 87.00 $\pm$ 0.37 & \cellcolor[rgb]{0.99, 0.73, 0.50} 94.15 $\pm$ 0.34 & \cellcolor[rgb]{0.99, 0.57, 0.26} 85.69 $\pm$ 1.03 & \cellcolor[rgb]{0.99, 0.56, 0.25} 93.73 $\pm$ 0.23\\ 
    & Fixed-lin. 2048 & \cellcolor[rgb]{0.95, 0.44, 0.11} 87.96 $\pm$ 0.28 & \cellcolor[rgb]{1.00, 0.96, 0.92} 93.10 $\pm$ 1.37 & \cellcolor[rgb]{0.95, 0.44, 0.11} 85.92 $\pm$ 1.15 & \cellcolor[rgb]{1.00, 0.96, 0.92} 91.99 $\pm$ 1.18\\ 
    & Distill 2048 & \cellcolor[rgb]{1.00, 0.90, 0.80} 86.85 $\pm$ 0.13 & \cellcolor[rgb]{0.95, 0.44, 0.11} 95.01 $\pm$ 0.07 & \cellcolor[rgb]{1.00, 0.96, 0.92} 84.73 $\pm$ 0.59 & \cellcolor[rgb]{0.96, 0.47, 0.14} 94.03 $\pm$ 0.10\\ 
    & Distill & \cellcolor[rgb]{1.00, 0.96, 0.92} 86.52 $\pm$ 0.05 & \cellcolor[rgb]{0.96, 0.46, 0.13} 94.97 $\pm$ 0.20 & \cellcolor[rgb]{1.00, 0.92, 0.84} 84.91 $\pm$ 0.19 & \cellcolor[rgb]{0.95, 0.44, 0.11} 94.12 $\pm$ 0.20\\ 
    \hline
    \end{tabular}
\end{table*}

\subsection{Additional experiments}\label{sec:misc}
Besides the results presented in the previous sections, we ran a few more experiments detailed in the supplementary materials. Here we briefly summarize the main findings.

Appendix A.2 discusses the influence of the choice of characterizing score by comparing the performance of 1C-Sum and T1000. As it turns out, 1C-Sum is better suited to the task.

Appendix A.3 illustrates that a single collection can serve for several (and quite different) tasks by promoting different subsets of samples.

Appendix A.4 shows that transfer is severely impeded when the student is not capable of matching the latent space of the teacher (\eg{} not complex enough).

\section{Conclusion}\label{sec:conclusion}
% TODO:
% - unfiromalizing policy
% - adding additional criteria for the sampling of biased samples (residual loss)
% Stratification ?
% Weakness: IPQR. How to ?

In this paper, we tackled the challenging task of distilling a large teacher network into a smaller one (the student) in the absence of the original data. We proposed to leverage a collection of unlabeled samples which is supposed to contain ``relevant'' samples. We focused on image classification for which such data bank is likely to exist.

To fully take advantage of the available collection, we proposed to (i) bias the sampling to present more often data which appear relevant in the sense of some (out-of-distribution) characterizing score, and (ii) better exploit the learning signal via fixed-linear distillation. To control the former, we introduced a simple hyper-parameter (IQPR). Contrary to related works, we are able to focus on relevant samples without requiring (a small part of) the original data and the whole learning runs in a time comparable to what would be required to directly learn the student, were the target dataset available.

We illustrated that good performances could indeed be reached when either the collection contained relevant samples or, unrealistically, the original data itself (Section \ref{sec:main-results}). We observed that biasing the sampling could speed up, or even help, the learning when irrelevant data is part of the collection (Section \ref{sec:result-sampling}). Biasing to the point where many datapoints are ignored might result in suboptimal performances, however. As for the fixed-linear distillation, we showed it was called for in our setting where the original data is missing (or we are on a tight learning budget); using more information from the latent space significantly helps the learning.

This paper opens some avenues for future works. Firstly, the way the biasing mechanism is controlled could be improved. In particular, varying the biasing as learning progress (\ie{} scheduling), aiming for a pseudo-class balance, or skipping samples on which the loss is relatively low might improve or speed up the transfer. Taking even better advantage of the learning signal is another promising direction.

To conclude, let us note that the most critical aspect is to have access to relevant data. Building a larger collection might well prevail to building a better one or improving the distillation algorithm.

{\small
\bibliographystyle{bibstyle}
\bibliography{main}
}

\pagebreak
\appendix

\section{Additional results}

\subsection{Biased sampling and uniformity}
Table \ref{tab:uniformity} as a fuller version of Table 4 of the main paper and illustrates how IQPR baises the sample selection.

\begin{table}[t]
	\caption{Biased sampling related metrics: skip ratio is the percentage of samples which are never selected; uniformity is the empirical entropy of the selected sample distribution; irrelevant proportion is the percentage of samples from \collection{irrel} which are used at least once during training (based on transferring CIFAR 10 from DenseNet 121 to MobileNet). Cells marked with $\star$ have non-negligible standard deviation (see Supplementary for more details).  \label{tab:uniformity}}
	\centering
	\small
	%\begin{sc}
	\begin{tabular}{c||c|c|c}
		\hline
		& \multicolumn{3}{c}{Skip ratio} \\
		IQPR & \collection{Rel + irrel} & \collection{Ori + irrel} & \collection{Ori} \\
		\hline
		1  & 0.00 $\pm$ 0.00  & 0.00 $\pm$ 0.00   & 0.00 $\pm$ 0.00  \\
		5  & 3.53 $\pm$ 0.01  & 1.91 $\pm$ 0.01   & 0.37 $\pm$ 0.01  \\
		25 & 24.99 $\pm$ 0.15 & 40.09 $\pm$ 15.83 & 21.48 $\pm$ 7.36 \\ 
		\hline
		\hline
		& \multicolumn{3}{c}{Uniformity} \\
		IQRP & \collection{Rel + irrel} & \collection{Ori + irrel} & \collection{Ori} \\
		\hline
		1  & 1.00 $\pm$ 0.00 & 1.00 $\pm$ 0.00 &  1.00 $\pm$ 0.00 \\
		5  & 0.97 $\pm$ 0.00 & 0.96 $\pm$ 0.00 & 0.94 $\pm$ 0.00 \\
		25 & 0.95 $\pm$ 0.00 & 0.86 $\pm$ 0.07 & 0.60 $\pm$ 0.18 \\
		\hline
		\hline
		& \multicolumn{3}{c}{Irrelevant proportion} \\
		IQPR & \collection{Rel + irrel} & \collection{Ori + irrel} & \collection{Ori} \\
		\hline
		1  & 58.36 $\pm$ 0.02 & 86.32 $\pm$ 0.00  & - \\
		5  & 42.12 $\pm$ 0.01 & 56.65 $\pm$ 0.01  & - \\ 
		25 & 38.84 $\pm$ 9.46 & 17.32 $\pm$ 12.81 & - \\
		\hline
	\end{tabular}
	%\end{sc}
\end{table}

\subsection{Influence of the characterizing score}
All the previous experiments relied on the 1C-Sum score to bias the sampling. Table \ref{tab:misc-ood} revisits the transfer of CIFAR 10 from ResNet 50 to MobileNet with our method using the \collection{rel + irrel} collection to highlight the effect of the characterizing score.

\begin{table}[t]
	\caption{Comparison of characterizing score. SK stands for skip ratio, uni. for uniformity and IP for irrelevant proportion (based on transferring CIFAR 10 from ResNet 50 to MobileNet).}
	\label{tab:misc-ood}
	\centering
	\begin{tabular}{cc||c|c|c|c}
		\hline
		\multicolumn{2}{l||}{IQPR \vphantom{2} Score} & Accuracy ($\%$) & SK & Uni. & IP \\
		\hline
		\multirow{2}{*}{5} & 1C-Sum & 91.04 $\pm$ 0.15 & 7.04 & 0.96 & 51.57 \\ 
		& T1000 & 89.57 $\pm$ 0.58 & 6.85 & 0.91 & 48.39\\ 
		\hline
		\multirow{2}{*}{25} & 1C-Sum & 89.25 $\pm$ 0.46 & 35.69 & 0.85 & 32.63 \\ 
		& T1000 & 74.69 $\pm$ 1.08 & 75.76 & 0.55  & 7.00 \\ 
		\hline
	\end{tabular}
\end{table}

The T1000 score results in slightly worse accuracy than 1C-Sum on moderate bias (IQPR=5). The accuracy drops significantly when the bias increases, however. Interestingly, the score seems appropriate to detect the relevant samples. At IQPR=5, both scores skip about $7\%$ of the data and T1000 slightly better rejects irrelevant samples. When IQPR=25, T1000 focuses on only a quarter of the samples but those are mostly relevant ones. In both cases, however, the results is a much less uniform distribution of samples compared to 1C-Sum.

This paradoxical situation is due to the fact that the characterizing score has two impacts on the learning (when IQPR $> 1$).

On the one hand, there is a direct effect: a bad choice of characterizing score would put forward irrelevant samples. In such a case, any IQPR value greater than one would result in a worse outcome than uniform sampling, since the student would be presented more often with ``bad'' samples. 

On the other hand, the score distribution is important. Imagine the score is perfect with respect to a given collection, task and teacher (informally, all relevant samples of the collection for task score higher than irrelevant ones) and there is exactly one quarter of irrelevant samples. The second quarter (containing only relevant samples) will almost never be selected if its components are much closer to the irrelevant samples than to third quarter, for instance. 

T1000 distribution is not suited for our problem, even though it might ultimately be a good characterizing score. A different scheme for transforming the score into a sampling probability might be more appropriate for T1000.

\subsection{One collection to rule them all}
An advantage of having a large collection of data from many sources is that the same collection can be used for different tasks. Using the same protocol as for the other experiments, we transferred by fixed-linear distillation a ResNet 50 teacher learned on KMNIST into a MobileNet student using the \collection{rel + irrel} collection (IQPR=25). We reached an accuracy of $97.45 \% \pm 0.80$. We thus see that re-using a collection which performs well on CIFAR 10 leads also to close-to-teacher  ($98.85 \% \pm 0.01$, see Table 2 of the main paper) accuracy on an unrelated problem. This time, the proportion of what was considered as ``irrelevant'' samples for CIFAR 10 (MNIST, Fashion MNIST, SVHN) increases up to reaching $75.09 \% \pm 4.04$.
Such samples are much more relevant with respect to KMNIST, illustrating that the characterizing score is indeed able to focus on the most relevant datapoints. 
%\ref{tab:teacher-acc}

%\begin{table}[t]
%	\caption{Additional experiments. KMNIST is transferred with fixed-linear distillation from ResNet 50 to MobileNet (IQPR=25). TwoConvNet (see Table \ref{tab:twoconv-details}) is used as student taught by ResNet 50 on CIFAR 10 (IQPR=1). In both cases, the \collection{rel+ irrel} collection is used.}
%	\label{tab:misc}
%	\centering
%	\begin{tabular}{c||c|c}
%		\hline
%		\multirow{2}{*}{KMNIST} & Accuracy & Irrel. prop. \\
%		%\cline{2-3}
%		& 97.45 $\pm$ 0.80 & 75.09 $\pm$ 4.04 \\
%		\hline
%		\hline
%		\multirow{2}{*}{TwoConvNet} & Fixed-lin. & Distill \\
%		%\cline{2-3}
%		& 56.06 & 58.46 \\
%		\hline
%	\end{tabular}
%\end{table}

\subsection{Failing the latent mapping assumption}
When proposing the fixed-linear distillation, we assumed that some correspondence between the teacher's and student's latent space existed. This seems to be the case when transferring from ResNet 50/DenseNet 121 to MobileNet/ShuffleNet. When the architectures are widely different, the assumption might not hold, however.  

To test this, we used as student a network composed of two depthwise-separable convolutions followed by a traditional linear part. This network will be referred to as TwoConvNet and is detailed in the appendix (we do not expect the architectural details to bear much weight on the conclusions, however). Although the number of parameters of this network is of the same order as MobileNet's (\ie{} roughly two millions parameters), TwoConvNet is far from being as deep.
We learned a TwoConvNet by fixed-linear distillation and classical distillation, following the same protocol as for the other experiments. We set the IQPR to 1 as we only wanted to test the mapping assumption. We obtained an accuracy of $56.06$ by fixed-linear distillation and of $58.46$ by classical distillation. This suggests that fixed-linear distillation is inappropriate when the latent spaces from the student cannot emulate the teacher's.  Admittedly, classical distillation does not work well either, since the network is too shallow for CIFAR 10.

\begin{table}[t]
	\caption{TwoConvNet: architecture details. DC stands for depthwise convolution, PC for pointwise convolution, BN for batch-normalization, GAP for global average pooling, FC for fully connected.}
	\label{tab:twoconv-details}
	\centering
	\begin{tabular}{c|c|c}
		\hline
		Layer type & Output size & Details \\
		\hline
		DC & $109 \times 109 \times 63$ & {\small kernel $7\times 7$, stride 2} \\
		PC & $109 \times 109 \times 512$ & {\small kernel $1 \times 1$} \\
		BN & $109 \times 109 \times 512$ & \\
		ReLU & $109 \times 109 \times 512$  &  \\
		\hline
		DC & $53 \times 53 \times 1024$ & {\small kernel $5\times 5$, stride 2} \\
		PC & $53 \times 53 \times 2048$ & {\small kernel $1 \times 1$} \\
		BN & $53 \times 53 \times 2048$  & \\
		ReLU & $53 \times 53 \times 2048$  & \\
		\hline
		GAP & $2048$ & \\
		FC & $10$ & \\
		\hline

	\end{tabular}
	
\end{table}
\end{document}